\documentclass[journal]{IEEEtran}
\usepackage{cite}
\usepackage{support-caption}
\usepackage{subcaption}
\usepackage[utf8]{inputenc}
\usepackage{caption}
\usepackage{multirow}
\usepackage{blindtext}
\usepackage{tabularx}
\usepackage[table,xcdraw]{xcolor}
\usepackage{array}
\usepackage{soul}
\usepackage{color}
\usepackage{booktabs}
\usepackage{times}
\usepackage[ruled,linesnumbered]{algorithm2e}
\usepackage{algorithmic}
\usepackage{soul}
\usepackage{url}
\usepackage[hidelinks]{hyperref}
\usepackage{amsmath}
\usepackage{graphicx}
\usepackage{listings}
\usepackage{xspace}
\usepackage{fancyhdr}
\usepackage{graphics}
\usepackage{makecell}
\usepackage{csquotes}
\usepackage{xr}
\usepackage{amssymb}
\usepackage{tikz}
\usepackage{pgfplots}
\pgfplotsset{compat=1.16}
\usepackage{wrapfig}
\usepackage{placeins}
\usepackage{subcaption} 
\usepackage{newfloat}
\usepackage{enumitem}
\usepackage{ulem} 
\usepackage{balance}
\usepackage[T1]{fontenc}
\definecolor{green}{rgb}{0.0, 0.5, 0.0}
\definecolor{amethyst}{rgb}{0.6, 0.4, 0.8}
\usepackage{colortbl}

\definecolor{codegray}{gray}{0.95}
\definecolor{commentgreen}{rgb}{0,0.6,0}
\definecolor{keywordblue}{rgb}{0.2,0.2,0.8}
\definecolor{stringpurple}{rgb}{0.58,0,0.82}

\lstdefinestyle{pythonstyle}{
    language=Python,
    backgroundcolor=\color{codegray},
    commentstyle=\color{commentgreen}\itshape,
    keywordstyle=\color{keywordblue}\bfseries,
    stringstyle=\color{stringpurple},
    basicstyle=\ttfamily\small,
    breaklines=true,
    frame=tb,
    numbers=left,
    numberstyle=\tiny,
    numbersep=6pt,
    showstringspaces=false,
    tabsize=4,
    captionpos=b,
    xleftmargin=1em,
    framexleftmargin=1em
}

\begin{document}
\title{Improving Adversarial Robustness Through Adaptive Learning-Driven Multi-Teacher Knowledge Distillation}

\author{Hayat Ullah, Syed Muhammad Talha Zaidi,
Arslan Munir,~\IEEEmembership{Senior Member,~IEEE,}
\thanks{Hayat Ullah and Arslan Munir are with the Intelligent Systems, Computer Architecture, Analytics, and Security Laboratory (ISCAAS Lab), Department of Electrical Engineering and Computer Science, Florida Atlantic University, Boca Raton, FL 33431, USA (e-mail: hullah2024@fau.edu, arslanm@fau.edu).}
\thanks{Syed Muhammad Talha Zaidi is with the Intelligent Systems, Computer Architecture, Analytics, and Security Laboratory (ISCAAS Lab), Department of Computer Science, Kansas State University, Manhattan, KS 66506, USA (e-mail: tzaidi@ksu.edu).}
}
\maketitle

\begin{abstract}
Convolutional neural networks (CNNs) excel in computer vision but are susceptible to adversarial attacks, crafted perturbations designed to mislead predictions. Despite advances in adversarial training, a gap persists between model accuracy and robustness. To mitigate this issue, in this paper, we present a multi-teacher adversarial robustness distillation using an adaptive learning strategy. Specifically, our proposed method first trained multiple clones of a baseline CNN model using an adversarial training strategy on a pool of perturbed data acquired through different adversarial attacks. Once trained, these adversarially trained models are used as teacher models to supervise the learning of a student model on clean data using multi-teacher knowledge distillation. To ensure an effective robustness distillation, we design an adaptive learning strategy that controls the knowledge contribution of each model by assigning weights as per their prediction precision. Distilling knowledge from adversarially pre-trained teacher models not only enhances the learning capabilities of the student model but also empowers it with the capacity to withstand different adversarial attacks, despite having no exposure to adversarial data. To verify our claims, we extensively evaluated our proposed method on MNIST-Digits and Fashion-MNIST datasets across diverse experimental settings. The obtained results exhibit the efficacy of our multi-teacher adversarial distillation and adaptive learning strategy, enhancing CNNs' adversarial robustness against various adversarial attacks.
\href{https://github.com/iscaas/Adverserial-Multi-Teacher-knowledge-Distillation}{\textcolor{purple}{\texttt{https://github.com/iscaas/MTKD-AR}.}}

\end{abstract}
\IEEEpeerreviewmaketitle
\section{Introduction}
\IEEEPARstart{D}{espite} the accomplishments of convolutional neural networks (CNNs) in real-world scenarios \cite{he2016deep, girshick2015fast, amodei2016deep}, recent studies have unveiled their vulnerability to adversarial examples, inputs crafted with subtle perturbations to confound network predictions \cite{szegedy2013intriguing, goodfellow2015explaining}. The addition of slight adversarial noise can easily deceive classification CNNs, leading to misclassifications. Even minor imperceptible alterations to an image can significantly alter a model's predictions, causing a notable decline in the accuracy of convolutional neural networks (CNNs). Adversaries can generate adversarial examples for a target model by exploiting perturbation transferability across surrogate models, without needing access to the target model itself. This poses a serious threat to domains reliant on deep learning with security implications, including autonomous driving, surveillance systems, and medical imaging etc. Thus, prioritizing the enhancement of adversarial robustness in constructing trustworthy AI systems becomes crucial for averting potential unforeseen risks.\\
\indent Efforts to fortify models' adversarial robustness have resulted in various defense techniques against adversarial examples, encompassing input denoising \cite{guo2017countering, liao2018defense}, detection strategies \cite{ma2018characterizing, lee2018simple}, and certifying classifier robustness \cite{cohen2019certified}. Adversarial training (AT) stands out as a particularly promising approach, involving direct augmentation of the training set with adversarial examples \cite{goodfellow2015explaining, madry2017towards}. While research actively seeks improved AT methodologies \cite{qin2019adversarial, zhang2019theoretically, wang2019improving}, another avenue for enhancing CNN's adversarial robustness involves modifying inputs during inference through techniques such as noise removal \cite{meng2017magnet}, super-resolution \cite{mustafa2019image}, and JPEG compression \cite{das2017keeping} to mitigate perturbation effects. However, these methods can be evaded by strong attacks \cite{athalye2018obfuscated}. While strategies based on adversarial training effectively enhance model robustness through crafted adversarial examples, they often come with computational costs and potential trade-offs in clean image classification performance. In addition to data-based modifications, certain approaches seek to enhance model robustness by altering network architectures or constructing network ensembles. However, these methods may entail additional processes and lack the flexibility required for widespread model adoption.\\
\begin{figure}[t]
  \centering
   \includegraphics[width=1.0\linewidth]{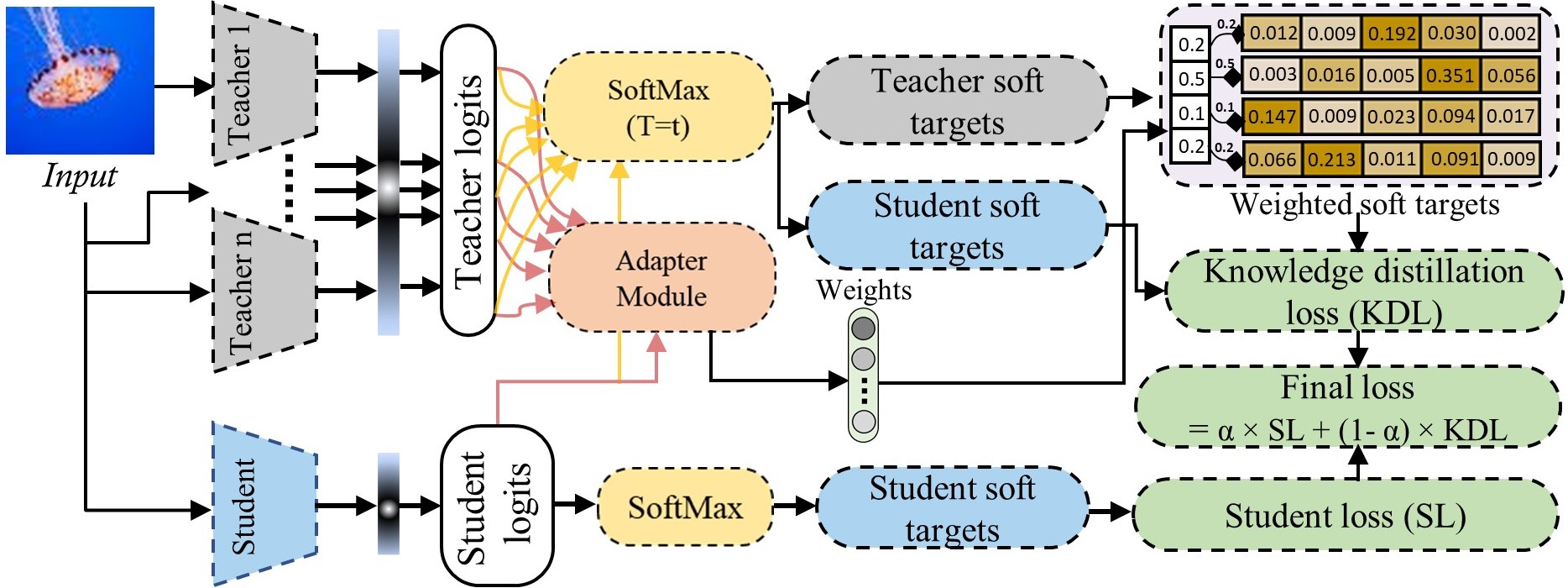}
   \caption{The brief visual overview and workflow of our proposed MTKD-AR method.}
   \label{fig:workflow}
\end{figure}
\indent Exploring strategies to enhance model resilience against adversarial attacks, it's important to note their inherent limitations. While effective, these methods often require high-capacity models for optimal results. Addressing this, knowledge distillation has emerged alongside adversarial training, aiming to bolster smaller deep neural networks (DNNs) by learning from larger, robust models \cite{goldblum2020adversarially, bai2021improving, chen2020robust, zhu2021reliable}. This approach, called Adversarial Robustness Distillation (ARD), adapts robustness techniques to smaller models, bridging the gap between their capacity and practical deployment needs. Despite these advancements, existing adversarial robustness techniques face several limitations. Adversarial training, while effective, incurs significant computational overhead due to the generation and augmentation of adversarial examples, making it difficult for resource-constrained environments. Moreover, adversarially trained models often exhibit limited generalizability to attack types beyond those seen during training, reducing their robustness in real-world scenarios. Ensemble-based approaches attempt to improve robustness by integrating predictions from multiple models, but they commonly assume equal contributions from all models, which can lead to suboptimal outcomes when individual models provide inaccurate or irrelevant guidance. To address these challenges, we propose the MTKD-AR framework, which employs dynamically weighted multi-teacher knowledge distillation to efficiently transfer adversarial robustness into a student model trained exclusively on clean data. This approach significantly reduces computational costs, enhances generalization across attack types, and avoids the pitfalls of static ensemble weighting strategies. More precisely, the main contributions of this work include:
\begin{enumerate}
    \item We propose a novel multi-teacher knowledge distillation framework, MTKD-AR, which enables the student model to achieve adversarial robustness without requiring adversarial data. The framework offers distillation from multiple adversarially trained teachers to a single student model, equipping the student to withstand diverse adversarial attacks.
    \item In multi-teacher knowledge distillation, not every teacher performs well for every single input image, teachers having inaccurate or irrelevant predictions can significantly influence and mislead the student model during the knowledge distillation process. To avoid this, we propose an adaptive learning technique that dynamically assigns importance weight to each teacher based on their prediction performance. This dynamic weight selection for each input image results in integrated predictions having higher importance to the prediction of the best-performing teacher. Multi-teacher knowledge distillation using our proposed adaptive learning technique ensures effective knowledge distillation.
    \item We conduct extensive quantitative experiments to demonstrate the effectiveness of our proposed method in terms of both knowledge distillation and adversarial robustness.
\end{enumerate}
The remainder of this paper is organized as follows. Section \ref{sec:RelatedWork} reviews the related work, outlining various methods developed for enhancing adversarial robustness in deep learning models. Section \ref{sec:problem_formulation} defines the problem formulation of adversarial attacks. The proposed MTKD-AR framework and its key components are detailed in Section \ref{sec:proposed_method}. Section \ref{sec:ExperimentalSettings} describes the experimental setup and implementation details. The obtained experimental results and comparisons with state-of-the-art methods are presented in Section \ref{sec:Results}. Finally, Section \ref{sec:Conclusion} concludes the paper and discusses potential directions for future research.
\section{Related Work}
\label{sec:RelatedWork}
An array of techniques has surfaced for crafting adversarial samples to exploit vulnerabilities in deep neural networks. The Fast Gradient Sign Method (FGSM), pioneered by \cite{szegedy2013intriguing}, efficiently generates adversarial samples by introducing noise aligned with the gradient's sign. It employs perturbations along the gradient direction of the cost function, using an epsilon parameter. Operating as a white-box attack, FGSM strategically manipulates inputs to produce subtle adversarial examples. The Refined Fast Gradient Sign Method (RFGSM) \cite{tramer2017ensemble}, an evolution of FGSM, introduces a slight random perturbation before linearizing the loss, yielding enhanced attack potency. In contrast to uniform initialization, \cite{xu2020exploring} employs a normal distribution with zero mean and variance in classical RFGSM. It also omits perturbation clipping post-FGSM, which leads to significant performance gains. The Basic Iterative Method (BIM) \cite{kurakin2018adversarial} advances FGSM by iteratively applying it to an image, elevating its attack effectiveness through multiple iterations and smaller step updates. Sharing similarities with BIM, the Projected Gradient Descent (PGD) method \cite{madry2017towards} enhances attack potency by incorporating a random starting point near the ground-truth image and employing a specified \textit{l}$_{1}$ norm-ball constraint. DeepFool \cite{moosavi2016deepfool} strategically adds minimal norm adversarial perturbations around decision boundaries to prompt false predictions. The Carlini and Wagner (C$\&$W) attack, designed by \cite{carlini2017towards}, utilizes an optimization-based strategy with a binary search mechanism, determining the minimal perturbation required for successful adversarial attacks. Additionally, C$\&$W seeks to balance imperceptible perturbations and attack strength through \textit{l}$_{0}$, \textit{l}$_{2}$, and \textit{l}$_{\infty}$ norm regularizations. Simultaneous Perturbation Stochastic Approximation (SPSA), a gradient-free method by \cite{uesato2018adversarial}, approximates gradients to generate effective adversarial attacks, excelling particularly in challenging optimization scenarios.\\
\indent To enhance model robustness and mitigate adversarial attacks, various defensive techniques have been proposed. One approach, rooted in adversarial training \cite{goodfellow2015explaining}, augments the training dataset with adversarial samples, showcasing effectiveness against diverse adversarial attacks. Further advancements in adversarial training have been explored, encompassing broader and larger models \cite{wu2021wider}, additional unlabeled data \cite{carmon2019unlabeled}, domain adaptation \cite{song2018improving}, and trade-offs between robustness and accuracy (known as TRADES) \cite{zhang2019theoretically} including the use of Kullback–Leibler divergence loss (KL loss). Other techniques involve emphasizing misclassified examples through Misclassification-Aware adversarial Training (MART) \cite{wang2019improving}, channel-wise activation suppression (CAS) \cite{bai2021improving}, and adversarial weight perturbation \cite{wu2020adversarial}. Common elements contributing to robustness in these methods involve the adoption of larger models, more data, and the incorporation of the KL loss.\\
\indent Adversarial training methods offer robustness improvements yet come with limitations. Smaller models like ResNet-18 and MobileNetV2 struggle to benefit from them \cite{gowal2020uncovering}, unlike larger counterparts such as WideResNet-34-10 and WideResNet-70-16. These methods are computationally expensive \cite{zhang2019theoretically} for adversarial sample generation, restricting their utility in resource-constrained scenarios like mobile devices, autonomous vehicles, and drones. To alleviate the computational burden, \cite{shafahi2019adversarial} proposed an efficient adversarial training algorithm that updates model parameters and image perturbations in a single backward pass. Beyond augmenting the training dataset, an alternative defense involves acquiring adversarially robust feature representations through model ensembles or network architecture alterations \cite{taghanaki2019kernelized, mustafa2019image, tramer2017ensemble, pang2019improving}.

\section{Problem Formulation} \label{sec:problem_formulation}
Adversarial attacks can be interpreted as the generation of small yet effective perturbations (modifications in input data) with the intention of manipulating the model’s behavior in a manner that might lead to incorrect or unexpected output.  More specifically, in the case of CNNs, which are commonly used for solving computer vision-related tasks (i.e., image classification, object detection, and image segmentation), perturbations refer to meticulously crafted modifications applied to the pixel values of input images. These alternations (pixel adjustments) are typically derived from the gradients of the model's loss function with respect to the input image. The main objective is to determine subtle pixel adjustments that, when added to the input image, can cause the model to misclassify the perturbed image, yet maintain the visual similarity to the original image.\\
\indent Considering the classification context, $f(x,\theta) : \mathbb{R}^{h \times w \times c} \rightarrow \{1, \ldots, k\}$ as an image classifier that maps an input image $x$ with a discrete label set $C$ containing $k$ classes. Here, $\theta$ represents the learnable parameters of $f$, while $h$, $w$, and $c$ denote the image's height, width, and channel dimensions, respectively. Given a perturbation magnitude $\epsilon$, the adversary aims to identify a perturbation $\delta \in \mathbb{R}^{h \times w \times c}$ to maximize the loss function, such as the cross-entropy loss ($l_{ce}$), in a manner that $f(x + \delta) \neq f(x)$. Further, $\delta$ can be derived as follows: 
\begin{equation}
\delta = argmax_{|\delta|_{p\leq\epsilon}} \mathcal{L}_{ce}(\theta, x + \delta, y)
\label{Eq:equation_1}
\end{equation}
\indent Here, $y$ is the true label (ground truth) of the input image $x$, where the term $|\delta|_{p}$ denotes the $L_{p}$ norm of the perturbation vector $\delta$, determining the magnitude of perturbation. The choice of $p$ determines the type of norm being employed. For instance, when $p = 2$, it corresponds to the $L_{2}$ or Euclidean norm, while $p = 1$ corresponds to the $L_{1}$ or Manhattan norm. Upon obtaining $\delta$, the perturbed image $x'$ can be derived as follows:
\begin{equation}
x' = x + \delta
\label{Eq:equation_2}
\end{equation}
\indent In response to these adversarial attacks towards CNNs, several defense strategies are proposed to mitigate the damage of adversarial attacks. These defensive approaches are developed considering the behavior of adversarial attacks to strengthen the model's efficiency to withstand adversarial attacks, ensuring the perseverance of the model's performance even in the presence of perturbed data. To better understand the defense mechanism against adversarial attacks, let's consider a CNN model $f$ (mapping function) having learnable parameters $\theta$ that learn how to map input data $x$ to predicted labels $y$. Generically, adversarial defense mechanisms can be formulated as an optimization problem, aiming to find the optimal parameters $\theta'$ that minimize the model's vulnerability to adversarial attacks. This optimization problem can be mathematically expressed as follows:
\begin{equation}
min_{\theta'} \mathcal{L}(\theta', x, y) + \lambda \cdot R(\theta')
\label{Eq:equation_3}
\end{equation}\\
\indent Here $L(\theta', x, y)$ is the loss function that estimates the model's performance on the clean input data $x$ with its corresponding label $y$. The variable $\lambda$ is the regularization parameter that balances the trade-off between the model's defense mechanism and its original performance. While $R(\theta')$ is the regularization term that encourages the model to have desirable properties that make it less susceptible to adversarial attacks. The choice of $R(\theta')$ can vary based on the particular defense strategy under consideration.

\section{Proposed Method} 
\label{sec:proposed_method}
To alleviate the impact of such attacks and strengthen the resilience of CNNs, we present a multi-teacher knowledge distillation (defensive distillation) approach, leveraging an adaptive learning mechanism that dynamically assigns importance weights to each teacher based on their prediction performance. While adversarial training and knowledge distillation techniques have been explored individually, our work introduces a novel integration of these approaches through a multi-teacher framework. The adaptive learning mechanism ensures context-sensitive knowledge transfer by prioritizing reliable teacher predictions. Unlike traditional adversarial training that requires adversarial data during the learning process, our method achieves robustness using clean data, significantly reducing computational costs and improving scalability.
\subsection{Adversarial Training Settings}
The primary aim of this study is to enhance the resilience of a CNN model against adversarial attacks without requiring exposure to perturbed data. To accomplish this objective, our approach entails training the model utilizing an adversarial training strategy. This strategy involves training the model to withstand a range of adversarial attacks, which encompass the fast gradient sign method (FGSM), fast feature fool gradient sign method (FFGSM), random feature gradient sign method (RFGSM), and projected gradient descent (PGD) attacks. To better understand, we interpret the adversarial training defensive technique as a min-max problem as given in Eq.  \ref{Eq:equation_4}. 
\begin{equation}
\min_{\theta} \frac{1}{|B|} \sum_{x,y \in B} \max_{\| \delta \| \leq \epsilon} \mathcal{L}(h_{\theta}(x + \delta), y)
\label{Eq:equation_4}
\end{equation}
\begin{figure*}[t]
  \centering
   \includegraphics[width=1.0\linewidth]{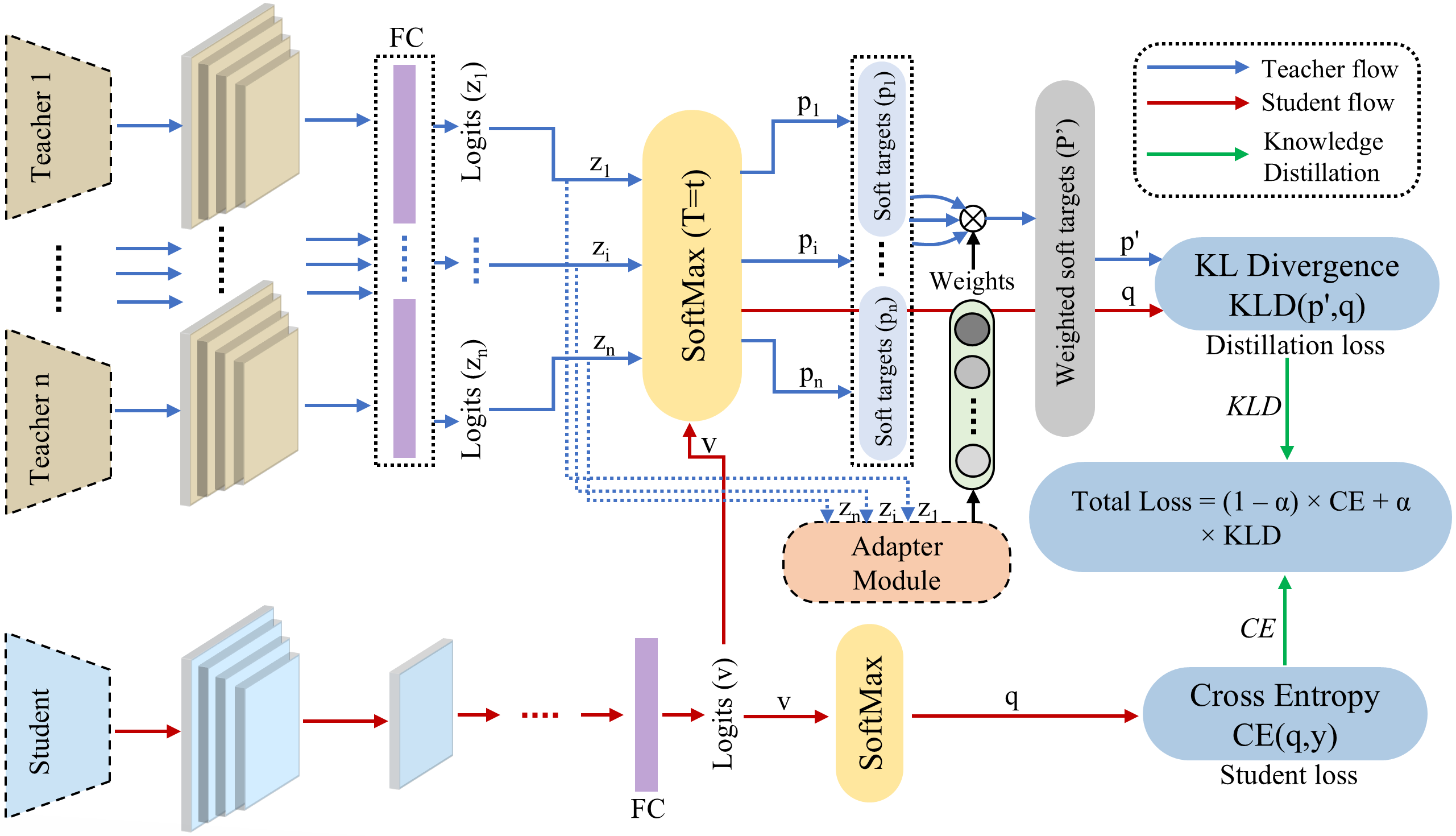}
   \caption{Detailed graphical overview of our proposed framework, depicting the overall workflow of adaptive learning-driven multi-teacher knowledge distillation for improving model robustness against adversarial attacks.}
   \label{fig:framework}
\end{figure*}
\indent Within this equation, we encounter two distinct objectives, comprising the inner maximization and the outer minimization components. The adversary strategically leverages the model's susceptibilities by introducing a perturbation $\delta$ with a magnitude of $\epsilon$ to the image $x$ from batch $B$ of image data. This manipulation steers the gradient flow of the model in the opposing direction and optimally maximizes the loss through the inner maximization process. In contrast, the outer minimization operation has the opposite role, minimizing the loss across multiple batches in the presence of perturbed data. In the context of the adversarial training formulation represented by Eq. \ref{Eq:equation_4}, the most straightforward approach for solving the outer-minimization problem and training an adversarially robust model involves the generation of perturbed image data using a defined adversarial attack. Subsequently, the generated adversarial data can be incorporated in the training process, which strengthens the model’s robustness through exposure to adversarially perturbed data during its learning process.\\
\begin{algorithm}[t]
\caption{Adversarial Training Algorithm}
\label{alg:algorithm}
\textbf{Input}: Training Dataset $D$\\
\textbf{Repeat}: \\
\qquad 1: Select minibatch $B$ from dataset $D$, initialize gradient vector g.\\
\qquad 2: For each (x,y) in $B$:\\
\qquad \qquad a. Find an attack perturbation $\delta^{*}$ by (approximately) optimizing\\ 
\qquad \qquad \qquad $\delta^{\star}(x) = \text{argmax}_{\| \delta \| \leq \epsilon} \mathcal{L}(h_{\theta}(x + \delta), y)$\\ 
\qquad \qquad b. Add gradient at $\delta^{*}$\\
\qquad \qquad \qquad $g:= g + \nabla_{\theta} \mathcal{L}(h_{\theta}(x + \delta^{*}(x)), y)$\\
\qquad 3: Update parameters $\theta$\\
\qquad \qquad \qquad $\theta := \theta - \frac{\alpha}{|B|} \cdot g$
\end{algorithm}
\indent To solve this outer minimization problem (known as adversarial training), let's reformulate Eq. \ref{Eq:equation_4} in the expression of the gradient descent function as given in Eq. \ref{Eq:equation_5} below:
\begin{equation}
\theta := \theta - \alpha \frac{1}{|B|} \sum_{x,y \in B} \nabla_{\theta} \max_{\| \delta \| \leq \epsilon} \mathcal{L}(h_{\theta}(x + \delta), y)
\label{Eq:equation_5}
\end{equation}
\indent Considering the mathematical property called "sub-differential of the maximum" or "sub-gradient of the maximum," we can rewrite the inner maximization objective as follows:
\begin{equation}
\nabla_{\theta} \max_{\| \delta \| \leq \epsilon} \mathcal{L}(h_{\theta}(x + \delta), y) = \nabla_{\theta} \mathcal{L}(h_{\theta}(x + \delta^{*}(x)), y)
\label{Eq:equation_6}
\end{equation}
Where,\\
\begin{equation}
\delta^{\star}(x) = \text{argmax}_{\| \delta \| \leq \epsilon} \mathcal{L}(h_{\theta}(x + \delta), y)
\label{Eq:equation_7}
\end{equation}
\indent Since $\delta^{*}(x)$ represents the value of $\delta$ that maximizes the $\mathcal{L}(h_{\theta}(x + \delta), y)$ as per magnitude constraints $\| \delta \| \leq \epsilon$, likewise, when the perturbation $\delta$ is adjusted in the permissible range of $\| \delta \| \leq \epsilon$ the associated loss $\mathcal{L}(h_{\theta}(x + \delta), y)$ takes its maximum value, resulting in $\delta^{*}(x)$. Thus, we can optimize the network parameters $\theta$ by computing the gradients with respect to $\delta^{*}(x) = \max_{\| \delta \| \leq \epsilon} \mathcal{L}(h_{\theta}(x + \delta), y)$. More precisely, the reformulated parameter optimization function can be expressed as follows:
\begin{equation}
\theta := \theta - \alpha \frac{1}{|B|} \times \nabla_{\theta} \mathcal{L}(h_{\theta}(x + \delta^{*}(x)), y)
\label{Eq:equation_8}
\end{equation}
\indent The working procedure of adversarial training is illustrated in Algorithm \ref{alg:algorithm}.
\subsection{Adversarial Robustness Distillation}
This section provides a comprehensive outline of our MTKD-AR technique, which distills knowledge from multiple adversarially trained models to a single student model. Based on the concept of learning robustness from diverse teacher models, our MTKD-AR approach equips the student model with the ability to withstand various adversarial attacks without any exposure to perturbed data. Figure \ref{fig:framework} illustrates the method's graphical overview, highlighting key components.
\begin{figure}[t]
  \centering
   \includegraphics[width=1.0\linewidth]{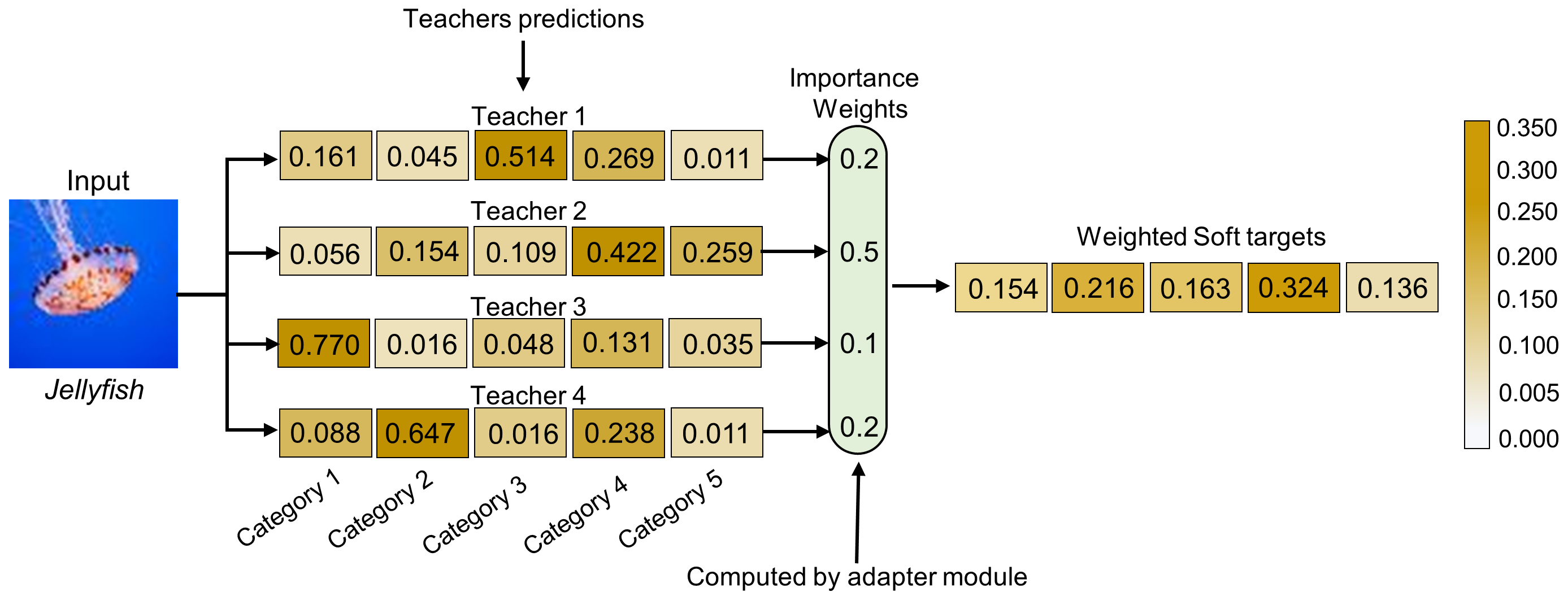}
   \caption{Pictorial overview of the proposed adaptive learning mechanism for multi-teacher knowledge distillation.}
   \label{fig:adavptive_learning}
\end{figure}
\subsubsection{Dynamic Weights Selection for Adaptive Learning}
The \textit{Adapter module} provides the foundation for the student model to learn from multiple adversarially trained teacher models. Typically, in the multi-teacher knowledge distillation paradigm, a straightforward approach involves aggregating the predictions from multiple teachers by computing their respective averages, where all teachers contribute equally to the final prediction, regardless of their individual performance. This naive approach, referred to as average learning, can result in less effective knowledge transfer when some teachers provide more accurate or valuable information than others. Additionally, inaccurate or irrelevant predictions from certain teachers can mislead the student model during the distillation process. To overcome these limitations, we propose an adaptive learning approach that dynamically assigns importance weights to each teacher based on their prediction performance, as depicted in Figure~\ref{fig:adavptive_learning}. The proposed adaptive learning mechanism uses cosine similarity to estimate the alignment between the logits of the student and teacher models.\\
\indent To better understand the proposed adaptive learning approach, let's consider multiple adversarially trained teacher models $T_{M_{(1,n)}}$ and a single student model $S_{M}$. During knowledge distillation, the adversarially trained teachers $T_{M_{(1,n)}}$ and student model $S_{M}$ produce logits (latent representation of input image from fully connected layer) represented by $z_{(1,n)}$ and $v$, respectively. The obtained logits $z_{(1,n)}$ and $v$ are then operates by a temperature-controlled softmax function which produces soft targets $P_{(1,n)}$ and $q$ for teachers and student, respectively. Mathematically, these operations can be expressed as follows:
\begin{equation}
P_{(1,n)} = \sum_{i=1}^{N} SoftMax(\overbrace{T_{M_{i}}\theta(x)}^{z_{(1,n)}})
\label{Eq:equation_9}
\end{equation}
\begin{equation}
q = SoftMax(\overbrace{S_{M}\theta(x)}^{v})
\label{Eq:equation_10}
\end{equation}
\indent Here, $T_{M_{i}}\theta$ represents $i_{th}$ teacher and $S_{M}\theta$ represents student model, where $x$ denotes the input image. To compute and assign a dynamic weightage to each teacher model as per their predictions, we compute the similarity between the logits $v$ and $z_{(1,n)}$, the obtained similarity between teachers and students is then used to compute a weight (importance vector) for a group of teacher models. Once computed, the obtained weights are then normalized to clip the weight values between [$0-1$]. The mathematical formulations for dynamic weight computation are given below in Eq. \ref{Eq:equation_11}, \ref{Eq:equation_12}, and \ref{Eq:equation_13}.
\begin{equation}
t_{similarity_{(1,n)}}= \sum_{i=1}^{N} Cosine Similarity (v, z_{i})
\label{Eq:equation_11}
\end{equation}
\begin{equation}
t^{\theta}_{(1,n)}= \sum_{i=1}^{N} 1 + t_{similarity_{i}}
\label{Eq:equation_12}
\end{equation}
\begin{equation}
t^{\theta'}_{(1,n)}= \sum_{i=1}^{N} \frac{t^{\theta}_{i}}{t^{\theta}_{1} + t^{\theta}_{2} + ... + t^{\theta}_{n} }
\label{Eq:equation_13}
\end{equation}
\indent Here, $v$ and $z_{i}$ are the student logits and $i_{th}$ teachers logits, respectively. $t_{similarity_{1,n}}$ is the similarity vector containing the similarity between student and teacher logits. The variable $t^{\theta}_{1,n}$ represents weights vector of teacher models, where $t^{\theta'}_{1,n}$ is normalized weights vector. The detailed pseudo-code of dynamic weight selection is given in Listing \ref{alg:teacher_weights}.\\
\indent The adaptive learning mechanism represents the core innovation of our framework. Unlike conventional ensemble techniques that treat all teacher models equally, our approach dynamically adjusts the contribution of each teacher based on its reliability for a given input. By computing cosine similarity between the logits of the student and teacher models, the framework ensures that the most reliable teachers are prioritized, minimizing the risk of misleading or irrelevant knowledge transfer.
\subsubsection{Multi-Teacher Knowledge Distillation with Adaptive Learning}
We propose an Adaptive Learning Driven multi-teacher knowledge distillation framework that supervises the student $\mathcal{S}$ model by leveraging the strengths of multiple adversarially trained teacher models $\{\mathcal{T}_{1}, \mathcal{T}_{2}, ..., \mathcal{T}_{n} \mid n = 4\}$. Our method enables the student model to effectively learn the implicit adversarial robustness of these teacher models without any direct exposure to adversarially perturbed data. To achieve this, we introduce a novel adapter module within the distillation process that dynamically assigns weights $\mathcal{W}_{i}$ to the teacher $\mathcal{T}_{i}$ model's predictions. This adaptive weighting scheme allows the student model to benefit from the collective knowledge of the teachers, guiding its learning process in a way that emphasizes robust feature learning. Consequently, our approach ensures that the student model develops inherent adversarial robustness, even when trained on clean data. The adaptive distillation algorithm not only refines the supervised knowledge provided to the student model but also enhances its generalization capabilities against adversarial attacks.
\begin{figure}[t]
\begin{lstlisting}[style=pythonstyle, caption={Pseudocod for Teacher Importance Weights Calculation.}, label={alg:teacher_weights}]
import numpy as np

def calculate_cosine_similarity(v, z):
    similarity = cosine_similarity(v, z)
    return similarity

# Calculate weights based on cosine similarity scores
def calculate_weights(similarity_score):
    # Map similarity scores to weights using a transformation
    weights = 1 + similarity_score
    return weights

# Normalize weights to ensure they sum up to 1
def normalize_weights(t_w):
    normalized_weights = []
    total_weights = np.sum(t_w)
    for i in range(len(t_w)):
        normalized_weights.append(t_w[i] / total_weights)
    return normalized_weights

# Calculate cosine similarity scores
teachers_similarities = []
for i in range(len(z)):
    teachers_similarities.append(calculate_cosine_similarity(v, z[i]))

# Obtaining importance weights of teachers
teachers_weights = []
for i in range(len(teachers_similarities)):
    teachers_weights.append(calculate_weights(teachers_similarities[i]))

# Normalize weights
normalized_teachers_weights = normalize_weights(np.array(teachers_weights))
\end{lstlisting}
\end{figure}
\indent During training, the distillation algorithm takes advantage of our proposed adapter module to dynamically assign weights to the prediction of each model as given in Listing \ref{alg:distillation}, providing the weighted prediction to better supervise the student model under training. The knowledge distillation loss in our adaptive learning settings can be formulated as follows: \\
\begin{equation}
distillation\;loss = \mathcal{KLD}(\overbrace{\sigma(z_{(1,n)} / \tau)}^{\overbrace{ p_{(1,n)} * W_{(1,n)}}^{p'}}, \overbrace{\sigma(v / \tau)}^{q}) \cdot \tau^2,
\label{Eq:equation_14}
\end{equation}
\indent Where $z_{(1,n)}$ and $v$ represent the logits (intermediate outputs) of the teacher and student models, respectively. These logits are processed through the sigmoid function, $\sigma$, resulting in the probability distributions $p_{(1,n)}$ for the teachers and $q$ for the student. The temperature parameter, $\tau$, controls the smoothness of the softmax outputs. The weight vector, $W_{(1,n)}$, assigns weights to the predictions of each teacher. The weighted soft targets of teacher models, denoted as $p'$, are obtained by computing the element-wise product of the teachers' probability vectors and their corresponding weights ($p_{(1,n)} * W_{(1,n)}$). 
\begin{figure}[t]
\begin{lstlisting}[style=pythonstyle, caption={Adaptive Learning-Driven Multi-Teacher Knowledge Distillation.}, label={alg:distillation}]
import tensorflow as tf

with tf.GradientTape() as tape:
    # Compute student's predictions
    student_predictions = student(_input, training=True)
    teachers_predictions = []

    for i in range(len(teachers_pool)):
        teachers_predictions.append(teachers_pool[i](_input, training=False))

    teachers_weighted_predictions = 0
    for i in range(len(normalized_teachers_weights)):
        teachers_weighted_predictions += (
            teachers_predictions[i] * normalized_teachers_weights[i]
        )

    # Compute knowledge distillation loss
    distillation_loss = compute_distillation_loss(
        student_predictions, teachers_weighted_predictions
    )

    # Compute student's own loss
    student_loss = student_loss_fn(student_predictions, labels)

    # Total loss with a balance parameter alpha
    total_loss = alpha * distillation_loss + \
                 (1 - alpha) * student_loss
\end{lstlisting}
\end{figure}
Further, the student model utilizes the standard Cross-Entropy (CE) loss as the key criterion for acquiring knowledge from the ground truth labels.
\begin{equation}
student\;loss = CE(y, q), \;\;where, \; q = \sigma(v)
\label{Eq:equation_15}
\end{equation}
\indent Here $v$ represents the logits of the student model after being processed by the sigmoid function $\sigma$, which yields $q$, the probability vector of the student model. While $y$ denotes the actual ground truth. Finally, the weighted sum of \textit{distillation loss} and \textit{student loss} is computed as follows:
\begin{equation}
total\;loss = (1 - \alpha) * student\;loss + \alpha * distillation\;loss, 
\label{Eq:equation_16}
\end{equation}
\indent The integration of dynamically weighted distillation loss with the standard supervised learning loss also represents a novel aspect of our framework. The tunable parameter $\alpha$ allows for precise control over the trade-off between clean-data accuracy and adversarial robustness, ensuring an optimal balance that has not been addressed in prior work.\\
\indent The design decisions in this work were guided by the objective of achieving robust adversarial defenses without requiring adversarial data for student training. By leveraging diverse adversarially trained teacher models and introducing an adaptive learning mechanism, the MTKD-AR framework ensures efficient knowledge transfer, scalable computational requirements, and enhanced generalization across diverse attack types. These innovations address key limitations in traditional adversarial training and ensemble learning methods.

\section{Experimental Settings}
\label{sec:ExperimentalSettings}
\subsection{Implementation Details}
The proposed MTKD-AR method is implemented in Python 3 and Tensorflow 2.0 Framework on a computing system with Intel (R) Xeon(R) CPU E5-2640 having a processor frequency of 2.50 GHz and 32 GB of dedicated main memory. The utilized computing system is equipped with two Tesla T4 GPUs having compute capabilities of 7.5 of 16 GB with Nvidia CUDA library version 12.0. The choice of the lightweight convolutional neural network (CNN) architecture was guided by its computational efficiency, making it suitable for deployment in resource-constrained environments. All scripts and implementation details have been shared in an online repository to facilitate reproducibility and verification of results.

\subsection{Datasets}
\subsubsection{MNIST-Digits}
This dataset is a collection of handwritten digits commonly used in computer vision-related tasks such as image classification and image reconstruction. It consists of 60,000 training images and 10,000 test images categorized in 10 classes (i.e., from digit 0 to 9). Each image is a single grayscale channel having a spatial resolution of $28\times28\times1$. In our experiments, we followed the standard protocol by using 60,000 images for training (in both adversarial and knowledge distillation training) and 10,000 images for model evaluation.

\subsubsection{MNIST-Fashion}
This dataset is similar to the original MNIST-Digits dataset, but instead of handwritten digits, it contains grayscale images of clothing items and accessories. It consists of 60,000 training images and 10,000 test images, each sized at $28\times28\times1$ pixels. The dataset includes 10 different categories of fashion items, such as shirts, shoes, dresses, etc. We followed the same training protocol for this dataset as the MNIST-Digits dataset, by utilizing 60,000 images for training and 10,000 images for model evaluation.   

\subsection{Training Details}
The proposed training scheme consists of two phases: the adversarial training phase and the multi-teacher knowledge distillation training phase. For adversarial training, we split the training data into clean training data and adversarial training data with a split ratio of 50$\%$. This approach allows the model to learn the discrimination between clean and adversarial images during training, minimizing the loss by optimizing the model parameters with respect to both clean and adversarial images. During adversarial training, we set the batch size to 64, and trained the network for 50 epochs using $Cross\:Entropy$ loss. To fine-tune the model parameters, we used $ADAM$ optimizer with a constant learning rate of $1e^{-3}$. For multi-teacher knowledge distillation training, we used only clean data, where the student model learns from the predictions of teacher models on clean data. This way, the student model not only learns the robustness in classification but also learns implicit characteristics of resilience from a group of teacher models to withstand adversarial attacks. For multi-teacher knowledge distillation training, we have used the same hyperparameters that are used in adversarial learning. 

\subsection{Threat Models}
In this study, we focus on four white-box attacks: FGSM, FFGSM, RFGSM, and PGD. White-box attacks were chosen as they represent the most challenging scenario, where the adversary has full access to model parameters and gradients. To systematically analyze the resilience of the student model, perturbation magnitudes (i.e., $\epsilon$ = 0.1, 0.15, 0.2 etc). were applied, reflecting real-world adversarial scenarios with increasing attack intensity. After attacking a model, we evaluate the attack transferability on a cloned version of the same network architecture with varying perturbation magnitudes. Further, models trained with a specific magnitude of perturbation are tested against higher magnitudes to assess their robustness under more aggressive adversarial conditions. The comparison between single-teacher and multi-teacher knowledge distillation demonstrates the advantage of leveraging diverse robustness characteristics from multiple adversarially trained models, showcasing the superior generalization and resilience of the MTKD-AR framework.

\section{Main Results}
\label{sec:Results}
\subsection{Comparison with Adversarially Trained Models}
To dive deeper into the comparative analysis, we evaluated the performance of the proposed MTKD-AR method against adversarially trained models. For a fair comparison, student models were trained under the supervision of adversarially trained teacher models ($\epsilon$ = 0.1). Once distilled, the performance of the distilled student was compared with that of the adversarially trained teachers. The results, presented in Table \ref{Table: MNIST_D_single_adv} and Table \ref{Table: MNIST_F_single_adv}, demonstrate that MTKD-AR consistently outperforms other comparative methods, even though it has no direct exposure to perturbed data during training. This validates the effectiveness of the MTKD-AR framework in achieving robust performance under both training-time and test-time adversarial attacks.\\
\indent The results also highlight the limitations of single-teacher adversarially trained models, which perform well only on the specific adversarial attacks they were trained for but suffer significant performance drops when evaluated on other types of attacks. In contrast, MTKD-AR achieves consistently high accuracy across all evaluated attack types, demonstrating its ability to generalize to unseen perturbations. By leveraging the collective robustness of multiple teachers, MTKD-AR ensures resilience to diverse adversarial scenarios while maintaining competitive performance on clean data. These findings underscore the effectiveness of the adaptive learning mechanism, which prioritizes knowledge transfer from the most reliable teachers for each input. Collectively, these observations validate the superiority of MTKD-AR over traditional adversarially trained models as a robust and generalizable defense mechanism.
\begin{table}[t]
    \centering
    \caption{\small Comparative Analysis of the Proposed Multi-Teacher Knowledge Distillation with Adversarial Training (MTKD-AR) Models: Evaluation on Clean and Perturbed Test Data of the MNIST-Digits Dataset with a Perturbation Budget of ($\epsilon$ = 0.1). The numbers highlighted as \colorbox{green!20}{best} and \colorbox{orange!20}{runner-up} indicate the top and second-best performances, respectively.}
    \resizebox{\columnwidth}{!}{%
    \begin{tabular}{c|c|c|c|c|c}
        \hline 
        \multicolumn{6}{c}{\textbf{MNIST-Digits Dataset Results}} \\ \hline
        Method & Clean data & FGSM & FFGSM & RFGSM & PGD \\ \hline
        CNN$_{WOKD}$ & \cellcolor{green!20}99.05 & 21.92 & 40.40 & 11.66 & 12.25 \\
        CNN$_{ADV-FGSM}$ & \cellcolor{orange!20}99.03 & \cellcolor{green!20}99.03 & $-$ & $-$ & $-$ \\
        CNN$_{ADV-FFGSM}$ & 96.89 & $-$ & \cellcolor{orange!20}91.32 & $-$ & $-$ \\ 
        CNN$_{ADV-RFGSM}$ & 97.43 & $-$& $-$ & \cellcolor{orange!20}93.29 & $-$ \\ 
        CNN$_{ADV-PGD}$ & 94.11 & $-$ & $-$ & $-$ & \cellcolor{orange!20}90.94 \\ 
        \textbf{Proposed (MTKD-AR)} & 97.66 & \cellcolor{orange!20}95.55 & \cellcolor{green!20}93.10 & \cellcolor{green!20}93.76 & \cellcolor{green!20}92.98 \\ \hline
\end{tabular}}
 \vspace{1 mm}
  \label{Table: MNIST_D_single_adv}
\end{table} 
\begin{table}[t]
    \centering
    \caption{\small Comparative Analysis of the Proposed Multi-Teacher Knowledge Distillation with Adversarial Training (MTKD-AR) Models: Evaluation on Clean and Perturbed Test Data of the MNIST-Fashion Dataset with a Perturbation Budget of ($\epsilon$ = 0.1). The numbers highlighted as \colorbox{green!20}{best} and \colorbox{orange!20}{runner-up} indicate the top and second-best performances, respectively.}
    \resizebox{\columnwidth}{!}{%
    \begin{tabular}{c|c|c|c|c|c}
        \hline
        \multicolumn{6}{c}{\textbf{MNIST-Digits Dataset Results}} \\
        \hline
        Method & Clean data & FGSM & FFGSM & RFGSM & PGD \\
        \hline
        CNN$_{WOKD}$ & \cellcolor{orange!20}91.41 & 17.91 & 28.53 & 13.47 & 15.30 \\
        CNN$_{ADV-FGSM}$ & \cellcolor{green!20}92.09 & \cellcolor{green!20}97.34 & $-$ & $-$ & $-$ \\
        CNN$_{ADV-FFGSM}$ & 91.18 & $-$ & \cellcolor{orange!20}87.48 & $-$ & $-$ \\ 
        CNN$_{ADV-RFGSM}$ & 88.68 & $-$& $-$ & \cellcolor{orange!20}86.56 & $-$ \\ 
        CNN$_{ADV-PGD}$ & 87.99 & $-$ & $-$ & $-$ & \cellcolor{orange!20}85.91 \\ 
        \textbf{Proposed (MTKD-AR)} & 91.19 & \cellcolor{orange!20}90.69 & \cellcolor{green!20}90.50 & \cellcolor{green!20}90.74 & \cellcolor{green!20}89.92 \\ \hline
\end{tabular}}
 \vspace{-6mm}
  \label{Table: MNIST_F_single_adv}
\end{table}
\subsubsection{Results on MNIST-Digits}
The results on the MNIST-Digits dataset, as illustrated in Table \ref{Table: MNIST_D_single}, offer comprehensive insights into the comparative performance of various models under different adversarial attack scenarios. The baseline model, denoted as $CNN_{WOKD}$, demonstrates commendable accuracy on clean data but exhibits significant vulnerability to adversarially perturbed inputs, underscoring the need for enhanced adversarial robustness. Leveraging knowledge distillation, the $CNN_{WKD-1T}$ model, trained with a single teacher, shows improved robustness against the specific attack it was trained on but struggles to generalize when faced with other adversarial perturbations. This highlights the challenge of achieving comprehensive robustness using a single-teacher paradigm.\\
\indent In contrast, the proposed MTKD-AR framework transcends these limitations, as demonstrated in Table \ref{Table: MNIST_D_single}. MTKD-AR consistently outperforms single-teacher models across all evaluated adversarial scenarios, maintaining high accuracy even under high perturbation magnitudes. Its dynamic weighting mechanism allows the model to effectively prioritize knowledge from the most reliable teachers, enabling superior generalization capabilities. These results reaffirm the effectiveness of adaptive learning-driven multi-teacher knowledge distillation in bridging the gap between clean-data accuracy and adversarial robustness. The findings underscore the advantage of leveraging diverse robustness characteristics from multiple teachers to achieve a more resilient defense mechanism than traditional single-teacher approaches.
\begin{table}[t]
    \centering
    \caption{\small Comparative Analysis of Results: Evaluating the results of Multi-Teacher Knowledge Distillation (MTKD-AR) in Contrast to Single Teacher Knowledge Distillation (WKD-1T) under Diverse Adversarial Attacks on the MNIST-Digits Dataset with a Perturbation Budget of ($\epsilon$ = 0.1). The numbers highlighted as \colorbox{green!20}{best} and \colorbox{orange!20}{runner-up} indicate the top and second-best performances, respectively.}
    \resizebox{\columnwidth}{!}{%
    \begin{tabular}{c|c|c|c|c|c}
        \hline
        \multicolumn{6}{c}{\textbf{MNIST-Digits Dataset Results}} \\
        \hline
        Method & Clean data & FGSM  & FFGSM & RFGSM & PGD \\
        \hline
        CNN$_{WOKD}$ & \cellcolor{green!20}98.89 & 21.92 & 40.40 & 11.66 & 12.25 \\
        CNN$_{WKD\_1T (FGSM)}$ & 97.22 & \cellcolor{green!20}95.79 & 51.16 & 48.36 & 52.79 \\
        CNN$_{WKD\_1T (FFGSM)}$ & 96.89 & 39.67 & \cellcolor{orange!20}91.32 & 44.53 & 49.41 \\ 
        CNN$_{WKD\_1T (RFGSM)}$ & 97.43 & 44.60 & 49.00 & \cellcolor{orange!20}93.29 & 47.53 \\ 
        CNN$_{WKD\_1T (PGD)}$ & 94.11 & 40.91 & 36.67 & 41.95 & \cellcolor{orange!20}90.94 \\ 
        \textbf{Proposed (MTKD-AR)} & \cellcolor{orange!20}97.66 & \cellcolor{orange!20}95.55 & \cellcolor{green!20}93.10 & \cellcolor{green!20}93.76 & \cellcolor{green!20}92.98 \\ \hline
\end{tabular}}
  \label{Table: MNIST_D_single}
\end{table}
\begin{table}[t]
    \centering
    \caption{\small Comparative Analysis of Results: Evaluating the Impact of Multi-Teacher Knowledge Distillation (MTKD-AR) in Contrast to Single Teacher Knowledge Distillation (WKD-1T) under Diverse Adversarial Attacks on the MNIST-Fashion Dataset with a Perturbation Budget of ($\epsilon$ = 0.1). The numbers highlighted as \colorbox{green!20}{best} and \colorbox{orange!20}{runner-up} indicate the top and second-best performances, respectively.}
    \resizebox{\columnwidth}{!}{%
    \begin{tabular}{c|c|c|c|c|c}
        \hline
        \multicolumn{6}{c}{\textbf{MNIST-Fashion Dataset Results}} \\
        \hline
        Method & Clean data & FGSM & FFGSM & RFGSM & PGD \\
        \hline
        CNN$_{WOKD}$ & \cellcolor{green!20}91.41 & 17.91 & 28.53 & 13.47 & 15.30 \\
        CNN$_{WKD\_1T (FGSM)}$ & 88.45 & \cellcolor{orange!20}87.16 & 41.36 & 51.90 & 47.13 \\
        CNN$_{WKD\_1T (FFGSM)}$ & 89.92 & 44.59 & \cellcolor{orange!20}89.54 & 40.18 & 51.39 \\ 
        CNN$_{WKD\_1T (RFGSM)}$ & 90.59 & 34.11 & 48.93 & \cellcolor{orange!20}88.47 & 49.10 \\ 
        CNN$_{WKD\_1T (PGD)}$ & 88.19 & 37.43 & 46.71 & 55.03 & \cellcolor{orange!20}87.94 \\ 
        \textbf{Proposed (MTKD-AR)} & \cellcolor{orange!20}91.19 & \cellcolor{green!20}90.69 & \cellcolor{green!20}90.50 & \cellcolor{green!20}90.74 & \cellcolor{green!20}89.82 \\ \hline
 \end{tabular}}
  \label{Table: MNIST_F_single}
\end{table}
\begin{table*}[t]
    \centering
    \caption{\small Comparative evaluation of quantitative results for our proposed approach on the MNIST-Digits dataset under varied adversarial attacks with different perturbation magnitudes ($\epsilon$ = 0, 0.1, 0.2, and 0.3). The numbers highlighted as \colorbox{green!20}{best} and \colorbox{orange!20}{runner-up} indicate the top and second-best performances, respectively.}
    \begin{tabular}{c|c|c|c|c|c}
        \hline
        \multirow{2}{*}{Method} & \multicolumn{5}{c}{Accuracy ($\%$)} \\
        \cline{2-6}
          & Clean ($\epsilon$ = 0) & FGSM ($\epsilon$ = 0.1) & FFGSM ($\epsilon$ = 0.1) & RFGSM ($\epsilon$ = 0.1) & PGD ($\epsilon$ = 0.1) \\
        \hline
        CNN$_{WOKD}$ & \cellcolor{orange!20}97.66 & 21.92 & 40.40 & 11.66 & 12.25 \\
        CNN$_{WKD\_1T (FGSM)}$ & 97.22 & \cellcolor{orange!20}95.55 & 51.16 & 48.36 & 52.79 \\
        CNN$_{WKD\_1T (FFGSM)}$ & 96.89 & 39.67 & \cellcolor{orange!20}91.32 & 44.53 & 49.41 \\
        CNN$_{WKD\_1T (RFGSM)}$ & 97.43 & 44.60 & 49.00 & \cellcolor{orange!20}93.29 & 47.53 \\ 
        CNN$_{WKD\_1T (PGD)}$ & 94.11 & 40.91 & 36.67 & 41.95 & \cellcolor{orange!20}90.94 \\
        \textbf{Proposed (MTKD-ADR)} & \cellcolor{green!20}98.89 & \cellcolor{green!20}95.79 & \cellcolor{green!20}93.10 & 
        \cellcolor{green!20}93.76 & \cellcolor{green!20}92.98 \\ \hline
          & Clean ($\epsilon$ = 0) & FGSM ($\epsilon$ = 0.2) & FFGSM ($\epsilon$ = 0.2) & RFGSM ($\epsilon$ = 0.2) & PGD ($\epsilon$ = 0.2) \\ \hline
        CNN$_{WOKD}$ & \cellcolor{orange!20}97.66 & 19.46 & 34.52 & 9.13 & 10.71 \\
        CNN$_{WKD\_1T (FGSM)}$ & 97.22 & \cellcolor{orange!20}93.74 & 48.39 & 44.82 & 43.40 \\
        CNN$_{WKD\_1T (FFGSM)}$ & 96.89 & 36.57 & \cellcolor{orange!20}89.63 & 41.53 & 46.09 \\ 
        CNN$_{WKD\_1T (RFGSM)}$ &97.43 & 42.81 & 45.16 & \cellcolor{orange!20}90.51 & 44.13\\ 
        CNN$_{WKD\_1T (PGD)}$ & 94.11 & 37.36 & 33.15 & 37.39 & \cellcolor{orange!20}88.69 \\ 
        \textbf{Proposed (MTKD-ADR)} & \cellcolor{green!20}98.89 & \cellcolor{green!20}94.13 & \cellcolor{green!20}92.12 & \cellcolor{green!20}91.51 & \cellcolor{green!20}90.06 \\ \hline
        & Clean ($\epsilon$ = 0) & FGSM ($\epsilon$ = 0.3) & FFGSM ($\epsilon$ = 0.3) & RFGSM ($\epsilon$ = 0.3) & PGD ($\epsilon$ = 0.3) \\ \hline
        CNN$_{WOKD}$ & \cellcolor{orange!20}97.66 & 17.33 & 31.19 & 8.56 & 9.11 \\
        CNN$_{WKD\_1T (FGSM)}$ & 97.22 & \cellcolor{orange!20}89.28 & 47.45 & 50.74 & 41.58 \\
        CNN$_{WKD\_1T (FFGSM)}$ & 96.89 & 35.14 & \cellcolor{orange!20}87.94 & 39.89 & 41.63 \\
        CNN$_{WKD\_1T (RFGSM)}$ & 97.43 & 40.57 & 39.12 & \cellcolor{orange!20}88.41 & 39.27 \\
        CNN$_{WKD\_1T (PGD)}$ & 94.11 & 35.00 & 29.67 & 35.56 & \cellcolor{orange!20}87.85 \\
        \textbf{Proposed (MTKD-ADR)} & \cellcolor{green!20}98.89 & \cellcolor{green!20}90.77 & \cellcolor{green!20}90.53 & \cellcolor{green!20}89.87 & \cellcolor{green!20}89.33 \\ \hline
    \end{tabular}
  \label{Table: MNIST_D}
\end{table*}
\begin{table*}[t]
    \centering
    \caption{\small Comparative evaluation of quantitative results for our proposed approach on the MNIST-Fashion dataset under varied adversarial attacks with different perturbation magnitudes ($\epsilon$ = 0, 0.1, 0.2, and 0.3). The numbers highlighted as \colorbox{green!20}{best} and \colorbox{orange!20}{runner-up} indicate the top and second-best performances, respectively.}
    \begin{tabular}{c|c|c|c|c|c}
        \hline
        \multirow{2}{*}{Method} & \multicolumn{5}{c}{Accuracy ($\%$)} \\
        \cline{2-6}
         & Clean ($\epsilon$ = 0) & FGSM ($\epsilon$ = 0.1) & FFGSM ($\epsilon$ = 0.1) & RFGSM ($\epsilon$ = 0.1) & PGD ($\epsilon$ = 0.1) \\
        \hline
        CNN$_{WOKD}$ & \cellcolor{orange!20}91.19 & 17.91 & 28.53 & 13.47 & 15.30 \\
        CNN$_{WKD\_1T (FGSM)}$ & 88.45 & \cellcolor{orange!20}87.16 & 41.36 & 51.90 & 47.13 \\
        CNN$_{WKD\_1T (FFGSM)}$ & 89.92 & 44.59 & \cellcolor{orange!20}89.54 & 40.18 & 51.39 \\ 
        CNN$_{WKD\_1T (RFGSM)}$ & 90.59 & 34.11 & 48.93 & \cellcolor{orange!20}88.47 & 49.10 \\ 
        CNN$_{WKD\_1T (PGD)}$ & 88.19 & 37.43 & 46.71 & 55.03 & \cellcolor{orange!20}88.94 \\
        \textbf{Proposed (MTKD-ADR)} & \cellcolor{green!20}91.91 & \cellcolor{green!20}90.69 & \cellcolor{green!20}90.50 & \cellcolor{green!20}90.74 & \cellcolor{green!20}90.82 \\ \hline
          & Clean ($\epsilon$ = 0) & FGSM ($\epsilon$ = 0.2) & FFGSM ($\epsilon$ = 0.2) & RFGSM ($\epsilon$ = 0.2) & PGD ($\epsilon$ = 0.2) \\ \hline
        CNN$_{WOKD}$ & \cellcolor{orange!20}91.19 & 14.62 & 25.49 & 10.55 & 12.45 \\
        CNN$_{WKD\_1T (FGSM)}$ & 88.45 & \cellcolor{orange!20}86.81 & 38.97 & 49.63 & 41.00 \\
        CNN$_{WKD\_1T (FFGSM)}$ & 89.92 & 42.08 & \cellcolor{orange!20}87.10 & 36.48 & 47.21 \\ 
        CNN$_{WKD\_1T (RFGSM)}$ &90.59 & 30.85 & 42.91 & \cellcolor{orange!20}87.15 & 45.62\\ 
        CNN$_{WKD\_1T (PGD)}$ & 88.19 & 33.44 & 42.64 & 49.28 & \cellcolor{orange!20}87.69 \\
        \textbf{Proposed (MTKD-ADR)} & \cellcolor{green!20}91.91 & \cellcolor{green!20}88.13 & \cellcolor{green!20}89.12 & \cellcolor{green!20}89.51 & \cellcolor{green!20}89.06 \\ \hline
        & Clean ($\epsilon$ = 0) & FGSM ($\epsilon$ = 0.3) & FFGSM ($\epsilon$ = 0.3) & RFGSM ($\epsilon$ = 0.3) & PGD ($\epsilon$ = 0.3) \\ \hline
        CNN$_{WOKD}$ & \cellcolor{orange!20}91.19 & 11.39 & 23.98 & 8.13 & 10.56 \\
        CNN$_{WKD\_1T (FGSM)}$ & 88.45 & \cellcolor{orange!20}83.51 & 34.65 & 45.91 & 37.25 \\
        CNN$_{WKD\_1T (FFGSM)}$ & 89.92 & 39.24 & \cellcolor{orange!20}85.45 & 32.52 & 40.31 \\
        CNN$_{WKD\_1T (RFGSM)}$ & 90.59 & 26.13 & 38.23 & \cellcolor{orange!20}85.14 & 39.28 \\
        CNN$_{WKD\_1T (PGD)}$ & 88.19 & 29.84 & 39.68 & 44.30 & \cellcolor{orange!20}86.85 \\
        \textbf{Proposed (MTKD-ADR)} & \cellcolor{green!20}91.91 & \cellcolor{green!20}86.15 & \cellcolor{green!20}87.64 & \cellcolor{green!20}86.11 & \cellcolor{green!20}88.47 \\ \hline
    \end{tabular}
  \label{Table: MNIST_F}
\end{table*}
\subsubsection{Results on MNIST-Fashion}
Building upon the insightful observations made in the context of the MNIST-Digits dataset, the examination of outcomes on the MNIST-Fashion dataset, as depicted in Table \ref{Table: MNIST_F_single}, provides a nuanced perspective on model robustness under varying adversarial attack conditions. The $CNN_{WOKD}$ baseline model, which exhibits commendable precision in classifying clean data, exposes susceptibility when exposed to adversarial distortions. This emphasizes the urgency of fortifying models against such perturbations. Notably, the $CNN_{WKD-1T}$ model, a product of single-teacher knowledge distillation, showcases improved resilience to the specific attack it was trained against, yet demonstrates diminished effectiveness against diverse adversarial manipulations. This outcome reaffirms the limitations of a solitary teacher-student paradigm in achieving comprehensive robustness. Remarkably, the novel multi-teacher knowledge distillation framework, MTKD-AR, surmounts these limitations by consistently showcasing resistance across a gamut of adversarial assaults. Impressively, MTKD-AR not only excels in adversarial datasets but also sustains impressive accuracy on clean data, reinforcing its capacity to harmonize accuracy and robustness.\\
\indent The results in Table \ref{Table: MNIST_F_single} highlight the efficacy of MTKD-AR on the MNIST-Fashion dataset. The $CNN_{WOKD}$ baseline model, while performing well on clean data, shows significant vulnerability under adversarial attacks. Similarly, single-teacher models achieve robust performance on the specific attacks they were trained to handle but fail to generalize to other perturbations. In contrast, MTKD-AR exhibits consistently high performance across all attack types, reflecting its ability to adapt to unseen adversarial scenarios. The combination of multiple teacher models and the adaptive learning mechanism enables MTKD-AR to overcome the limitations of single-teacher frameworks, ensuring robust and generalizable defense capabilities. Furthermore, the consistent performance of MTKD-AR on clean data highlights its ability to maintain classification accuracy without compromising robustness.
\subsection{Ablation Study}
\subsubsection{Impact of Different Epsilon Values on the Model's Performance}
This section presents additional results on the MNIST-Digits and MNIST-Fashion datasets, focusing on their perturbed versions under varying magnitudes of adversarial attacks ($\epsilon$ = 0.1, 0.2, 0.3). The obtained quantitative results, tabulated in Table \ref{Table: MNIST_D} and Table \ref{Table: MNIST_F}, compare the performance of the proposed MTKD-AR method against baseline and single-teacher models on both clean and perturbed data. The results demonstrate that MTKD-AR consistently outperforms other methods across all ($\epsilon$) levels, maintaining superior accuracy even as the perturbation strength increases. This robustness highlights the effectiveness of the adaptive learning mechanism in leveraging the collective robustness of multiple adversarially trained teachers to transfer adversarial resilience to the student model.\\
\indent Notably, MTKD-AR achieves competitive performance on clean data and significantly higher accuracy on adversarial datasets compared to single-teacher and baseline ensemble methods. Unlike traditional approaches, which rely on adversarially trained data for student training, MTKD-AR achieves resilience without exposing the student model to perturbed data during training. These findings emphasize the scalability and generalization capabilities of MTKD-AR, making it a reliable defense mechanism against adversarial perturbations of varying intensities.
The trends observed across both datasets further validate the robustness of MTKD-AR under challenging scenarios. On clean data ($\epsilon$ = 0), all models perform commendably, but with adversarial perturbations ($\epsilon$ = 0.1, 0.2, 0.3), single-teacher models exhibit notable accuracy degradation, particularly when tested on attacks they were not specifically trained for. In contrast, MTKD-AR demonstrates resilience across all evaluated conditions, underscoring its adaptability and ability to generalize effectively. These results reinforce the practicality of MTKD-AR in handling diverse adversarial scenarios, making it a robust choice for real-world applications.

\section{Conclusion and Future Work}
\label{sec:Conclusion}
In this paper, we introduce a novel approach for enhancing the robustness of convolutional neural networks (CNNs) through multi-teacher adversarial robustness distillation combined with an adaptive learning strategy. Our method involves training multiple adversarially trained clones of a lightweight CNN model using various adversarial attacks. These adversarially trained models then act as teachers, guiding the learning of a student model on clean data through a multi-teacher knowledge distillation process. To ensure effective robustness enhancement, we introduce an adaptive learning strategy that assigns weights based on the prediction precision of each model. By distilling knowledge from adversarially pre-trained teacher models, our approach not only strengthens the student model's learning capabilities but also equips it to withstand adversarial attacks, even when trained solely on clean data. Rigorous experimentation across diverse settings underscores the efficacy of our proposed method. The experimental outcomes provide compelling evidence for the potency of our multi-teacher adversarial distillation framework and the adaptive learning strategy in bolstering CNNs' adversarial robustness.\\
\indent The proposed MTKD-AR framework presents several opportunities for further exploration and development. Its scalability and adaptability make it highly applicable to large-scale deployments in domains where adversarial robustness is crucial, such as autonomous systems, medical imaging, and cybersecurity. By eliminating the need for adversarial data during student training, MTKD-AR reduces computational costs, making it ideal for resource-constrained environments, including mobile platforms, drones, and IoT devices. Future research could focus on extending the adaptive learning mechanism to other domains, such as natural language processing and transformer-based architectures, where robustness to adversarial attacks remains a significant challenge. Additionally, integrating MTKD-AR with certified defenses or federated learning setups could address pressing security and privacy concerns in distributed environments, ensuring its relevance to emerging AI systems. Automating the selection of teacher models and exploring the integration of fairness or domain adaptation techniques may further enhance the robustness, scalability, and versatility of the framework. These advancements would enable MTKD-AR to adapt to diverse data distributions and ensure equitable performance across different user demographics. By pursuing these directions, the framework has the potential to serve as a reliable and scalable solution for real-world adversarial challenges across various domains.



\bibliographystyle{IEEEtran}
\bibliography{References}
\vskip -2\baselineskip plus -1fil

\vskip -2\baselineskip plus -1fil

\end{document}